# Atrial scar segmentation via potential learning in the graph-cut framework

Lei Li[1, 2], Guang Yang[3], Fuping Wu[2], Tom Wong[3], Raad Mohiaddin[3], David Firmin[3], Jenny Keegan[3], Lingchao Xu[4], Xiahai Zhuang[2*]

[1] School of Biomedical Engineering, Shanghai Jiao Tong University, China
[2] School of Data Science, Fudan University, Shanghai, China
[3] National Heart and Lung Institute, Imperial College London, UK
[4] School of NAOCE, Shanghai Jiao Tong University, Shanghai, China

**Abstract.** Late Gadolinium Enhancement Magnetic Resonance Imaging (LGE MRI) emerges as a routine scan for patients with atrial fibrillation (AF). However, due to the low image quality automating the quantification and analysis of the atrial scars is challenging. In this study, we proposed a fully automated method based on the graph-cut framework, where the potential of the graph is learned on a surface mesh of the left atrium (LA), using an equidistant projection and a deep neural network (DNN). For validation, we employed 100 datasets with manual delineation. The results showed that the performance of the proposed method was improved and converged with respect to the increased size of training patches, which provide important features of the structural and texture information learned by the DNN. The segmentation could be further improved when the contribution from the t-link and n-link is balanced, thanks to the inter-relationship learned by the DNN for the graph-cut algorithm. Compared with the existing methods which mostly acquired an initialization from manual delineation of the LA or LA wall, our method is fully automated and has demonstrated great potentials in tackling this task. The accuracy of quantifying the LA scars using the proposed method was 0.822, and the Dice score was 0.566. The results are promising and the method can be useful in diagnosis and prognosis of AF.

## 1    Introduction

Atrial fibrillation (AF) is the most common arrhythmia of clinical significance, occurring in up to 2% of the population and rising fast with advancing age. It is associated with structural, contractile and electrical remodeling, and also related to increased morbidity and mortality. Catheter ablation (CA) using the pulmonary vein (PV) isolation technique has emerged as the most common methods for the AF patients who are not responding to pharmacological treatments.

Generally, the clinical reference standard technique for the assessment of atrial scars is the electro-anatomical mapping (EAM) [1], performed during an electrophysiological (EP) study prior to CA. However, due to its invasiveness, the use of ionizing radiation and suboptimal accuracy, a non-invasive Late Gadolinium Enhancement Magnetic

* Corresponding author: zxh@fudan.edu.cn. This work was supported by Science and Technology Commission of Shanghai Municipality (17JC1401600).

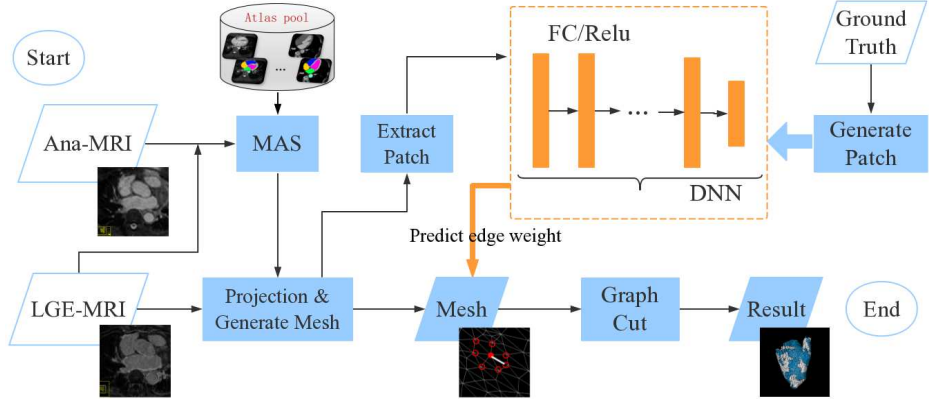

Fig. 1 Flowchart of the proposed segmentation framework.

Resonance Imaging (LGE MRI) is a promising alternative and potentially a more accurate imaging technique. LGE MRI allows the detection of native fibrosis and ablation induced scaring based on the mechanism of slow washout kinetics of the gadolinium in the scar area.

Before the delineation of atrial scars, an accurate segmentation of the atrium wall is required for the exclusion of outer structures with the similar intensities as the atrial scars and the inclusion of all the scar areas. However, automating this segmentation remains challenging due to the possible poor quality of the LGE MRI images, the thin thickness of the left atrium (LA) wall (~1-3mm) and the various LA shape. Most of the methods for scar segmentation used the manual segmentation of the LA or LA wall to provide an accurate initialization for scar delineation [2,3]. There is a benchmark paper which evaluated different methods for the LA segmentation [4]. Vein *et al.* [5] proposed an algorithm named ShapeCut, which combined a shape-based system and the graph cut approach to make a Bayesian surface estimation for the LA wall segmentation. Currently, the most widespread method for scar segmentation is still based on thresholding. However, the choice of an appropriate threshold relies on subjective opinions from a domain expert, eventually limiting the external validation and reproducibility of this method. A reliable and reproducible method to detect and segment atrial scars on the LA wall remains an open question.

In this paper, we present a novel fully automatic framework for the segmentation and quantification of atrial scars. We tackled the quantitative analysis of atrial scars by developing a graph-based segmentation with the learned t-/n-link potential in the flatten LA surface. It starts by the segmentation of the LA endocardium from whole heart segmentation (WHS) by combining LGE MRI and the anatomical 3D MRI, referred to as Ana-MRI based on b-SSFP sequence. Then the segmented LA wall is projected onto the surface for a graph cut. Finally, the two weights (n-link and t-link potentials) of the graph are designed and explicitly learned from a Deep Neural Work (DNN).

## 2 Method

Fig. 1 presents the overall workflow of the proposed method, which consists of three major components: (1) 2D graph-cut with well-defined n/t link weights was performed to automatically segment the atrial scars from the LA wall, (2) an equidistant projection of the 3D LA endocardium to a 2D map and (3) explicit learning of the potentials for edge weights of the graph.

### 2.1 Initialization of atrial endocardium

Multi-atlas segmentation (MAS) of whole heart was used to generate the initial segmentation of LA endocardium, which includes the multi-atlas propagation based on image registration and label fusion [6]. WHS offers adequate information of the LA boundary and there's a multi-modality WHS (MM-WHS) challenge which was held in conjunction with STACOM and MICCAI 2017 [7]. As the LGE MRI images could have relatively poor image quality, we propose to propagate the segmentation result of Ana-MRI to the LGE MRI using a primitive registration [8]. After the atlas propagation, we adopt the local weighted label fusion to cover the different texture patterns of LGE MRI and b-SSFP images. At the same time, a multi-scale multi-modality patch-based label fusion algorithm was used [9], because the intensity distribution of LA blood pool is almost the same as that of the blood pool from other chambers.

Having finished the WHS for LA and PV delineation (mean Dice = 0.89), we then generate an initial LA image labeled with blood pool and mitral valve. The classical marching cubes algorithm was then used to obtain a mesh surface of LA wall which excludes the mitral valve.

### 2.2 Graph formulation for scar segmentation via equidistant projection

Classification and quantification of scars can be formulated as a 2D graph-cut problem with well-defined n/t link weights. In our study, the graph is formed by a set of vertices and edges defined in a surface, which is obtained from the 3D LA endocardium by an equidistant projection (details in Section 2.3). It should be noted preservation of distance is required in the definition of n-link weights in the proposed graph cut framework. Associated with each vertex is an intensity profile that is consisted of a patch defined along the normal direction of the target surface. The cost function of the graph cut problem includes both region and boundary terms of the segments, which can be explicitly learned (details in Section 2.4).

Let $G = \{S, \mathcal{N}\}$ denote a graph, where $S = \{p_i\}$ indicates the set of graph nodes, $\mathcal{N} = \{<p_i, p_j>\}$ is the set of edges. Let $l_{p_i} \in \{0, 1\}$ be the label assigned to $p_i$ and $l = \{l_{p_i}, p_i \in S\}$ be the label vector that defines a segmentation. Then the segmentation energy can be defined as:

$$E(l|\theta^0, \theta^1) = E_D(l|\theta^{l_{p_i}}) + \lambda E_B(l)$$
$$= \sum_{p_i \in S} W_{p_i}^{t-link}(l_{p_i}) + \lambda \sum_{(p_i, p_j) \in \mathcal{N}} W_{\{p_i, p_j\}}^{n-link} \delta(l_{p_i}, l_{p_j}) \qquad (1)$$

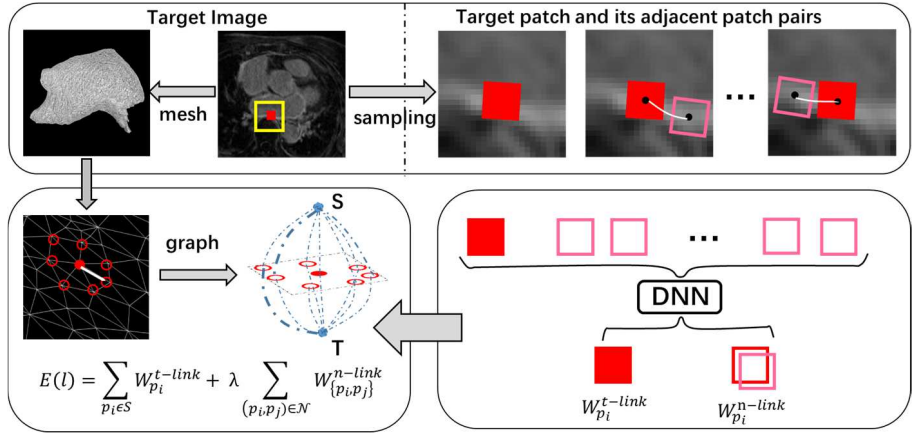

Fig. 2 Pipeline of our proposed method at the training and testing phases

where $\delta\left(l_{p_i}, l_{p_j}\right)$ is the Kronecker delta function, namely the function is 1 if the two variables are equal, and 0 otherwise.

In conventional graph-based segmentation [10], the appearance models ( $\theta^0, \theta^1$ ) are considered as a prior by manual defining some seed points (i.e., graph-cut) or giving a bounding box for interactive segmentation (i.e., grab-cut). Different from these interactive methods, we propose to directly learns the t- and n-link weights using a DNN.

### 2.3 Projection of the atrial myocardium

Localization and quantification of the atrial scars can be made using the EAM system, which only focuses on the surface area of the LA. Inspired by this, we proposed a method to allow simultaneous representation of multiple parameters on a surface based on the template of an average LA mesh that is similar as Williams et al. [11]. In this scheme, the atrial scar can be classified on the flattened surface projected from the 3D LA geometry, thereafter both the errors due to LA wall thickness and the misregistration of the WHS can be mitigated. In this formulation, each vertex (or node) on the surface should include an intensity profile that represents the texture information of corresponding location in the LGE MRI. This idea is incorporated into the learning of the graph representing the surface.

### 2.4 Explicit learning of graph potentials

The atrial scars are classified using the graph-cut algorithm on the projected surface graph, whose t-/n-link potentials are learned using a DNN. Fig. 2 provides the pipeline of the training and testing phases.

For the training samples, which are nodes on the graph with intensity profiles, we propose to extract a patch at the corresponding location in the LGE MRI by back projection. The local orientation of the patches is defined on the local coordinate of the pixel, such that the x- and y- axis of the patch are aligned with the tangent plane of the surface in the 3D geometry of the LA myocardium. To further mitigate the effect of misalignment of the LA surface, which is computed from the WHS, an internal and external extension along the normal direction of the surface is considered.

After the construction of patch library, the labeling of them is based on the ground truth of scar and the WHS. For training of the t-link potentials, patches with labels are considered. For training of the n-link, a sample is defined with a patch pair (associated with two connected nodes), the geodesic distance between the two nodes, and their similarity. Here, the similarity is computed from their labels or probability of being a label, with higher similarity value if they have the same label or higher probability of being the same label, and lower similarity otherwise.

To learn the t-link potentials, we use a fully connected neural network. Feeding the patch around a node into the framework, one can obtain the probability of the node belonging to scar or background separately. Similarly, we designed another neural network to calculate the n-link of two connected nodes, with the corresponding two patches and distance as the input. To account the distance, the training procedure begins by projecting each patch into a multi-feature space and then combines with the distance feature as a whole to obtain the final weight.

## 3      Experiment

### 3.1   Data Acquisition

We acquired 100 datasets for experiments. The patients were with longstanding persistent AF on a 1.5T Siemens Magnetom Avanto scanner (Siemens Medical Systems, Erlangen, Germany). Transverse navigator-gated 3D LGE-CMRI was performed using an inversion prepared segmented gradient echo sequence (TE/TR 2.2ms/5.2ms) 15 minutes after gadolinium administration. The inversion time was set to null the signal from normal myocardium. Detailed scanning parameters are 30–34 slices at (1.4–1.5)×(1.4–1.5)×4mm$^3$, reconstructed to 60–68 slices at (0.7–0.75)×(0.7–0.75)×2mm$^3$, field-of-view 380×380mm$^2$. For each patient, prior to contrast agent administration, coronal navigator-gated 3D b-SSFP (TE/TR 1ms/2.3ms) data were also acquired with the following parameters: 72–80 slices at (1.6–1.8)×(1.6–1.8)×3.2mm$^3$, reconstructed to 144–160 slices at (0.8–0.9)×(0.8–0.9)×1.6mm$^3$, field-of-view 380×380mm$^2$. Both LGE-CMRI and b-SSFP data were acquired during free-breathing using a crossed-pairs navigator positioned over the dome of the right hemidiaphragm with navigator acceptance window size of 5mm and CLAWS respiratory motion control.

### 3.2   Gold standard and evaluation

The 100 LGE MRI images were all manually segmented by experienced cardiologists specialized in cardiac MRI to label the enhanced atrial scar regions. The manual delineation is regarded as gold standard for experiments. The 100 datasets were randomly divided into two equal-size sets, one for training and the other for test. For the evaluation, we computed the accuracy, sensitivity, specificity and Dice scores between the classification sets, respectively from ground truth and the automatic segmentation.

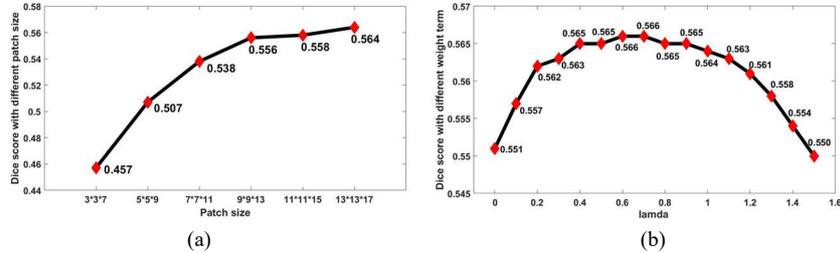

(a) (b)

Fig. 3: Dice scores of the proposed method with different parameterizations: (a) performance against different patch sizes; (b) performance against the balancing parameter λ to weight the t-link term and n-link term in the graph-cut framework.

Table 2. Summary of the quantitative evaluation methods.

| Method | Accuracy | Sensitivity | Specificity | Dice |
| --- | --- | --- | --- | --- |
| Ostu | 0.496±0.235 | 0.853±0.225 | 0.290±0.188 | 0.360±0.164 |
| GMM | 0.716±0.096 | 0.961±0.049 | 0.370±0.160 | 0.467±0.155 |
| LearnGC | 0.822±0.065 | 0.932±0.048 | 0.515±0.167 | 0.566±0.140 |
| LearnGC (post) | 0.807±0.070 | 0.921±0.046 | 0.552±0.149 | 0.652±0.084 |
| LearnGC (pre) | 0.839±0.054 | 0.987±0.049 | 0.360±0.135 | 0.465±0.125 |

We first investigated the influence of different patch sizes. Then, the performance of the proposed method with different value of the balancing parameter λ was studied. Finally, the optimal parameters were used for the comparison study. It should be noted that in the comparison study, the proposed method is equivalent to a classification method which is purely based on the DNN for prediction of scars when $\lambda = 0$.

## 4  Result and Discussion

Fig. 3 (a) presents the mean Dice scores of the proposed method using different sizes of the patches. One can see from the figure that the Dice coefficient first increases dramatically, then tends to converge after the patch size becoming more than $9 \times 9 \times 13$, and the best Dice score was obtained when the method used the largest size of patches. This is reasonable, as the larger size of the patches is used, the richer intensity profile is included for feature training and detection. When the patch size is larger than it is necessary, the DNN will simply discard the useless information in the training and detection, thanks to the well-designed architecture of the DNN.

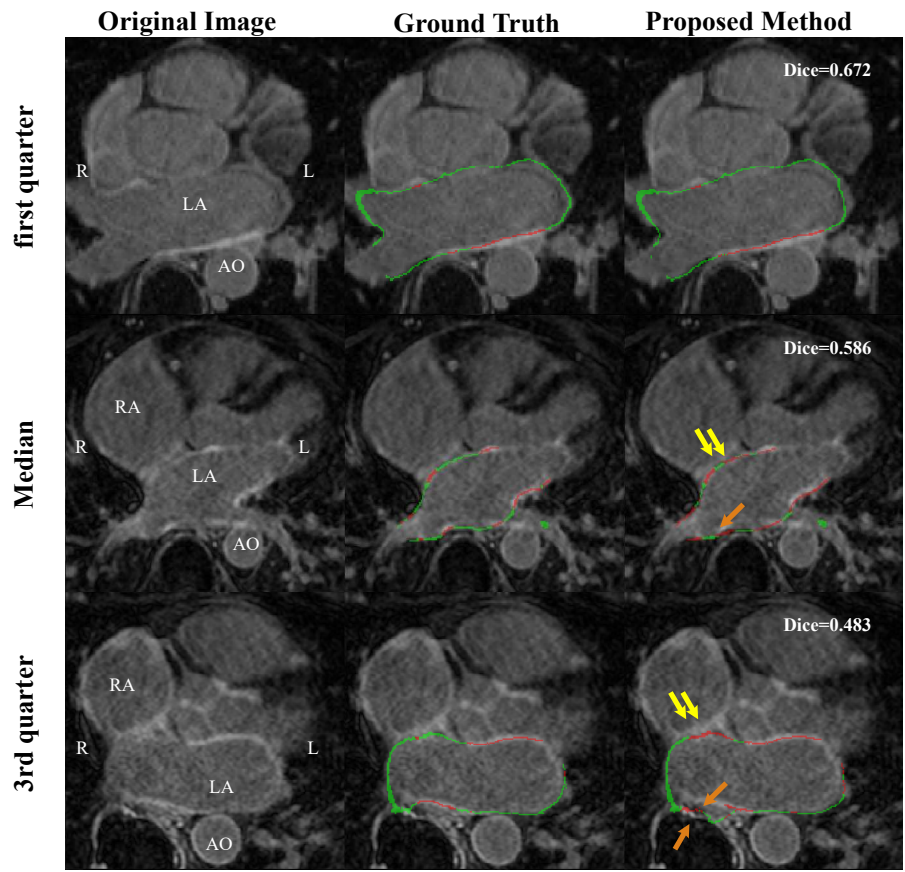

Fig.4: 2D visualization of the final atrial scar segmentation results of the first quarter (a), median (b) and third quarter cases (c) in the test dataset by the proposed method. Double yellow arrows in (b) and (c) show the enhancement due to the effect of the scar from right atrium (RA). And single orange arrows in (b) and (c) show the errors due to the misregistration of WHS.

The result of the parameter study on λ is presented in Fig. 3 (b). One can see that the best performance in terms of Dice score was obtained when λ is set to 0.6, meaning inter-node relation is important and should be considered for scar classification, and the weight between the t-link and n-link terms should be balanced, in order to achieve the optimal results.

For comparisons, we evaluated two conventional algorithms, i.e. the Ostu intensity threshold [12] and the Gaussian mixture model (GMM) [13] for intensity classification. Both methods required the initialization of the LA wall segmentation from the WHS of the corresponding Ana-MRI image in each dataset. The quantitative results of the two methods are presented in Table 1, compared with the proposed learning graph-cut algorithm, referred to as LearnGC in the table. By compared the performance of pre- and post-ablation scans separately, we conclude that the quantification of the pre-ablation cases is more challenging. It could be due to the fact that the fibrosis appears more diffuse and has greater overlap with normal wall. In addition, we implemented the threshold (UTA) and graph cuts (KCL), both benchmarked in the Challenge [3], but initialized by our automatic LA segmentation. Their Dice scores, UTA: 0.360 and KCL: 0.485, were both significantly lower than ours (0.566) ($p<0.0001$).

Fig. 4 visualizes three examples from the test set, i.e. the first quarter, median and third quarter cases in terms of Dice scores using the proposed method. One can see that the method could provide promising results for localization and quantification of atrial scars. In the median and third quarter cases, we highlighted the errors which are commonly seen in the segmentation results. First, the double arrows identify the potential erroneous classification due to the enhancement of neighbor tissues, such as the right atrial scars. At these locations where the left and right atrial walls are joint together, it can be arduous to differentiate the boundaries. The orange arrows show the WHS result had relatively large errors in delineating the endocardium, resulting in a boundary shift of the projecting of the classified scars. One may find that even the WHS had errors, the proposed method still can identify the scars at the corresponding locations of the projecting surface. This is mainly attributed to the effective training of the DNN, which assigns random shifts along the perpendicular direction of the surface when extracting the training patches. This capability contributes to less demanding of the WHS accuracy for providing initialization and reconstruction of the LA surface mesh.

## 5   Conclusion

We proposed a fully automated method for segmentation and quantification of atrial wall scars. Two major technical contributions are introduced. One is the graph-cut framework combined with the idea of projecting the 3D LA wall onto a surface graph for graph-based segmentation. The other is the adoption of the DNN, which extracts the features from local intensity profiles of LGE MRI for learning the relations or similarity of node pairs in the constructed graph. In addition, we achieved the automatic segmentation of LA, which can provide an accurate enough initialization for scar segmentation. The proposed method performs better when the size of extracted patches increases, but the performance converges when the size is larger than a certain value. The learning graph-cut method demonstrated evidently better results than the two conventional methods ($p<0.0001$). The Dice scores for quantifying LA and LA scars are respectively 0.891 and 0.566.